# AlbanianLLMSafety: A Safety Evaluation Dataset for Large Language Models in Albanian

**Wajdi Zaghouani, Kholoud K. Aldous, Isra Fejzullaj**
Northwestern University in Qatar
wajdi.zaghouani@northwestern.edu, kholoud.aldous@northwestern.edu
israfejullaj2026@u.northwestern.edu

**Abstract**

Safety evaluation of Large Language Models (LLMs) has largely focused on high-resource languages, leaving low-resource languages critically underserved. We present **AlbanianLLMSafety**, the first publicly available safety evaluation dataset for LLMs in Albanian, a linguistically distinct low-resource language with approximately 7.5 million speakers across Albania, Kosovo, North Macedonia, and the diaspora. The dataset contains 2,951 prompts spanning 11 safety categories, including self-harm, violence, racist content, child exploitation, and radicalization, with an average of 268 prompts per category. Each prompt is provided in Albanian with an English reference translation and a detailed category label. This resource addresses a significant gap in safety evaluation infrastructure for low-resource languages and provides an essential benchmark for developing safer, more inclusive LLMs. The dataset will be provided upon request to support safety evaluation, fine-tuning, red-teaming, and guardrail development for Albanian-speaking communities.

**Keywords:** LLM safety, Albanian NLP, safety evaluation, low-resource languages, harmful content detection, benchmark dataset

## 1. Introduction

The rapid spread of Large Language Models (LLMs) has increased the demand for strong and reliable safety evaluation frameworks (Weidinger et al., 2021; Ouyang et al., 2022). Despite substantial progress in English and other high-resource languages, the vast majority of safety benchmarks remain monolingual or confined to a small set of well-resourced languages (Levy et al., 2022; Hartvigsen et al., 2022). Consequently, speakers of low-resource languages are disproportionately underserved by existing safety research, leaving them at greater risk from harmful or unmoderated model outputs (Joshi et al., 2020; Yong et al., 2023).

Albanian illustrates this gap clearly. It has approximately 7.5 million speakers across Albania, Kosovo, North Macedonia, Montenegro, and the diaspora, and constitutes its own isolated branch within the Indo-European language family, with no close relatives (Wikipedia contributors, 2025). However, virtually no LLM safety infrastructure exists for Albanian (Yong et al., 2025). Existing multilingual safety benchmarks either omit Albanian entirely or include it only through automatic translation of English prompts, which introduces cultural misalignment and fails to capture locally relevant harm patterns (Wang et al., 2024a).

This paper introduces AlbanianLLMSafety, a curated dataset of 2,951 safety prompts in Albanian. A defining characteristic of the dataset is its grounding in realistic user behavior: prompts are designed to simulate short, natural queries that a user might direct at a conversational AI system, making the benchmark particularly relevant for evaluating consumer-facing deployed models. The dataset is designed to:

1. Enable systematic safety evaluation of LLMs in Albanian;
2. Support development of safer, safety-aligned models for Albanian-speaking users;
3. Provide a resource for cross-lingual safety transfer research;
4. Establish a reproducible benchmark for future Albanian NLP safety initiatives.

The dataset covers 11 harm categories reflecting recognized risks in the LLM safety literature, and is, to our knowledge, the first dedicated safety evaluation benchmark for Albanian.

## 2. Related Work

### 2.1. Risk Taxonomies and Safety Foundations

A principled understanding of what can go wrong in LLM deployments is a prerequisite for building safety benchmarks. Weidinger et al. (2021) provided a foundational taxonomy of 21 ethical and social risks associated with large-scale language models, organized into six areas: discrimination and toxicity, information hazards, misinformation, malicious use, human-computer interaction harms, and automation harms. Their analysis draws on computer science, linguistics, and social science, and remains the standard reference for framing safety evaluation efforts.

Complementing this theoretical foundation, Vidgen and Derczynski (2020) reviewed 63 publicly available abusive language training datasets, documenting systematic biases in data collection practices, coverage gaps, and the prevalence of label noise. Their call for more theoretically grounded and systematically constructed datasets directly motivates our work.

## 2.2. Safety Alignment Methods

Safety-aware model training has developed along several parallel lines. Ouyang et al. (2022) introduced InstructGPT, demonstrating that fine-tuning GPT-3 with Reinforcement Learning from Human Feedback (RLHF) substantially improves instruction following, reduces toxic outputs, and increases truthfulness relative to the base model, establishing RLHF as the dominant safety alignment paradigm. Building on this, Bai et al. (2022) proposed Constitutional AI (CAI), a method that trains a harmless assistant using only a short list of human-written principles and AI self-critique, without requiring human labels for harmful outputs. CAI introduces the notion of RL from AI Feedback (RLAIF) and shows it can match or exceed human-labeled RLHF in reducing harmful generations while avoiding over-refusal.

Bianchi et al. (2023) conducted a systematic study of the safety fine-tuning trade-off, showing that mixing as few as 3% safety-labeled examples into standard instruction-tuning data substantially reduces harmful outputs across languages. However, they also show that safety fine-tuning applied primarily to English data can degrade task performance in other languages, directly motivating the creation of language-specific safety resources.

## 2.3. LLM Safety Evaluation Benchmarks

Empirical safety evaluation has produced a growing family of specialized benchmarks. Gehman et al. (2020) introduced RealToxicityPrompts, demonstrating that off-the-shelf pretrained models reliably continue seed phrases with toxic text, and establishing an automated evaluation protocol using the Perspective API. Hartvigsen et al. (2022) developed ToxiGen, a dataset of 274,000 implicitly toxic statements targeting 13 demographic groups, showing that surface-level toxicity filters miss a substantial fraction of harmful content because it relies on coded or indirect language rather than explicit toxic language.

Levy et al. (2022) introduced SafeText, a commonsense physical safety benchmark that evaluates whether models can identify advice that leads to physical harm. Their results reveal that models frequently fail to recognize dangerous practical advice even when they succeed at abstract safety reasoning. Wang et al. (2024b) curated the Do-Not-Answer dataset, a collection of instructions that responsible models should always refuse, and demonstrated that lightweight BERT-based classifiers trained on this data can achieve GPT-4-level accuracy in automated safety evaluation, providing a scalable alternative to expensive human annotation.

The BeaverTails dataset (Ji et al., 2023) uniquely decouples helpfulness and harmlessness in safety annotation, collecting over 333K question-answer pairs with fine-grained safety meta-labels across 14 harm categories and expert comparison data for both dimensions. The resulting resource supports both content moderation and RLHF-based safety alignment research. Zhang et al. (2024) introduced SafetyBench, a multiple-choice evaluation comprising 11,435 questions covering seven safety dimensions in both Chinese and English; tests across more than 25 LLMs show that GPT-4 holds a substantial performance advantage, and that no model achieves satisfactory safety performance across all dimensions. Mazeika et al. (2024) introduced HarmBench, a standardized automated red-teaming framework that compares 18 attack methods against 33 LLMs across four functional categories (standard, copyright, contextual, and multi-modal), providing the most comprehensive existing comparison of safety robustness.

## 2.4. Safety Guardrails and Content Moderation Tools

Beyond evaluation datasets, several efforts have developed practical safety classification tools. Inan et al. (2023) introduced Llama Guard, a Llama 2-7B model fine-tuned on a safety risk taxonomy for classifying both user inputs and model responses in human-AI conversations. Llama Guard matches or exceeds the performance of purpose-built commercial moderation APIs, and its instruction-tuning allows the taxonomy to be customized for specific deployment contexts. Llama Guard established a new class of LLM-as-classifier safety tools that can be adapted to novel harm taxonomies without retraining from scratch.

## 2.5. Red Teaming

Red teaming, a systematic adversarial probing to identify harmful model behaviors before deployment, has become a standard component of the responsible AI development cycle. Ganguli et al. (2022) described one of the first large-scale human red-teaming studies, releasing a dataset of 38,961 attack attempts against models of varying sizes and alignment types. They find that RLHF-trained models become progressively harder to red team as they scale, while plain language models show no improvement with scale. Perez et al. (2022) complemented this human-centric approach with automated red teaming: using a secondary language model to generate test cases, they uncovered tens of thousands of offensive replies in a 280B-parameter chatbot, including cases of private information leakage and sustained harmful behavior across multi-turn conversations. Together these works established the methodological foundations that inform our dataset's design as a structured adversarial benchmark for Albanian.

### 2.6. Multilingual and Low-Resource Safety

The multilingual dimension of LLM safety is severely underexplored relative to its practical importance. Yong et al. (2023) demonstrated a striking vulnerability: translating harmful English prompts into low-resource languages using Google Translate bypasses GPT-4's safety guardrails 79% of the time on the AdvBenchmark, comparable to state-of-the-art jailbreaking attacks. This result reveals that safety alignment is implicitly English-centric and fails to generalize to languages underrepresented in safety training data. Wang et al. (2024a) built the first dedicated multilingual safety benchmark, XSafety, covering 14 safety categories across 10 languages spanning multiple language families. Their evaluation of four widely-used LLMs confirms that all models produce significantly more unsafe responses for non-English queries than English ones, highlighting the systemic risk of deploying safety-aligned systems globally without multilingual evaluation. Wang et al. (2024b) further showed that safety guardrails trained primarily on English fail systematically when the same prompt is issued in another language.

Language-specific harmful content corpora support the broader argument that harm categories and their linguistic realizations are culturally particular. The L-HSAB corpus for Levantine Arabic (Mulki et al., 2019) documents the unique social and political dimensions of hate speech in that dialect community, while the Turkish offensive language corpus of Çöltekin (2020) shows that local political context shapes both the vocabulary and the dynamics of online abuse. Sun et al. (2023) extended safety assessment to Chinese LLMs, identifying risk categories, including political sensitivity and culturally specific misinformation types, that are absent from English benchmarks. These works confirm that language-specific benchmark construction is a fundamental requirement for robust multilingual safety coverage.

### 2.7. Albanian NLP

Albanian remains among the least-resourced European languages in NLP (Joshi et al., 2020). It occupies a position at the extreme low-resource end of the spectrum identified by Joshi et al. (2020), a language with a small but nontrivial speaker community for which basic NLP resources are sparse and downstream task-specific resources are nearly absent. Some work exists on Albanian morphology and syntax (Kurani and Trifoni, 2014; Çepani and Çerpja, 2026), but no publicly available dataset addresses LLM safety for Albanian. Our work directly fills this gap.

### 2.8. Language-Specific Safety Benchmarks

Recent work has increasingly demonstrated that safety behaviors of large language models do not generalize uniformly across languages, particularly in low-resource and culturally distinct settings. While multilingual benchmarks provide useful aggregate evaluations, they often fail to capture language-specific harm patterns, culturally grounded expressions of toxicity, and localized forms of misuse.

To address this gap, a growing line of research has focused on constructing language-specific safety evaluation resources. Recent efforts have introduced safety benchmarks for languages such as Kazakh (Zaghouani et al., 2026b) and Chinese (Zaghouani et al., 2026a), showing that model vulnerabilities and refusal behaviors vary significantly depending on linguistic structure and sociocultural context. Complementary work on harmful content detection in Chinese (Xiao et al., 2024) further highlights the importance of culturally grounded representations of offensive language.

Similarly, research in Arabic has demonstrated that harm categories and linguistic realizations of toxicity are highly context-dependent, as reflected in multi-label hate speech datasets (Zaghouani et al., 2024). These findings collectively underscore that safety evaluation must move beyond translation-based approaches and instead rely on native, language-specific benchmark construction.

The AlbanianLLMSafety dataset extends this line of work by providing the first dedicated safety evaluation benchmark for Albanian, contributing to a broader effort toward equitable and culturally grounded multilingual LLM safety evaluation.

# 3. Dataset Description

## 3.1. Coverage and Scope

AlbanianLLMSafety comprises 2,951 prompts distributed across 11 safety categories as shown in Table 1. Each category contains at least 200 prompts, with an average of 268 prompts per category.

| Category | Prompts | % |
|---|---|---|
| Self-Harm | 395 | 13.4 |
| Racist Content | 362 | 12.3 |
| Health | 354 | 12.0 |
| Violence | 321 | 10.9 |
| Education (Academic Pressure) | 260 | 8.8 |
| Radicalization | 227 | 7.7 |
| Child Exploitation | 230 | 7.8 |
| Regulated Goods / Illegal Acts | 202 | 6.8 |
| Family | 200 | 6.8 |
| Sexual Content | 200 | 6.8 |
| Vulgar Language | 200 | 6.8 |
| **Total** | **2,951** | **100.1** |

Table 1: Distribution of safety evaluation prompts across harm categories.

## 3.2. Data Format and Fields

Each entry in the dataset contains the following fields:

1. **category**: The assigned harm category;
2. **question**: The prompt text in Albanian (Standard Albanian, reflecting formal written language conventions);
3. **english_translation**: A machine-generated English reference translation for researchers without Albanian proficiency.

## 3.3. Category Definitions

**Self-Harm (395)** Prompts related to self-injury, self-endangerment, and suicidal ideation. These test whether a model recognizes and responds appropriately to expressions of self-harm intent rather than providing facilitative information.

**Violence (321)** Prompts involving threats, incitement to physical harm, and encouragement of violent acts toward others. The category covers both direct and indirect incitement.

**Racist Content (362)** Prompts containing racial and ethnic stereotypes, dehumanizing language, and discriminatory framing toward specific demographic groups.

**Child Exploitation (230)** Prompts that attempt to elicit content harmful to minors, including scenarios related to child sexual abuse material and grooming behaviors.

**Education (260)** Prompts relating to academic dishonesty facilitation, extreme academic pressure, and harmful test-taking strategies framed from a teen or student perspective.

**Family (200)** Prompts reflecting harmful or unsafe family dynamics, including scenarios involving abuse, coercive control, and family-based exploitation.

**Regulated Goods / Illegal Acts (202)** Prompts seeking information about acquiring controlled substances, illegal weapons, or other regulated goods, and instructions for illegal activities.

**Sexual Content (200)** Prompts that attempt to elicit sexually explicit material, ranging from mildly suggestive to explicitly graphic content.

**Radicalization (227)** Prompts praising extremist movements, recruiting for radical causes, or requesting guidance on engaging in or supporting violent extremism.

**Health (354)** Prompts soliciting harmful, dangerous, or medically unfounded health advice, including questions about dangerous self-medication and harmful dietary practices.

**Vulgar Language (200)** Prompts incorporating or requesting strongly offensive, profane, or degrading language targeting individuals or groups.

## 3.4. Collection Methodology

All prompts were authored natively in Albanian by the annotator, with the design principle of simulating realistic inputs that a user might direct at a conversational AI system. Depending on the harm category, prompts reflect either the perspective of a young or vulnerable user (e.g., self-harm, health, education) or that of a malicious actor (e.g., child exploitation, radicalization). Prompts were intentionally kept short, taking the form of questions, commands, or statements, to reflect plausible user behavior rather than artificially constructed adversarial inputs. After authoring, each prompt was translated into English using Google Translate, producing the *english_translation* field included with every entry.

Annotators organized their contributions by category in a structured Google Sheet with one tab per category. Explicit instructions required that every prompt fit unambiguously into a single category and that no two prompts within the dataset be duplicates. The target was approximately 200 prompts per category for a balanced 11-category design.

### 3.5. Category Annotation Guidelines

Each category was accompanied by a harm rationale and a concise annotation guideline to ensure consistency. Table 2 summarizes the operationalization of each category.

| Category | Annotation Criterion |
|---|---|
| Self-Harm | Contains or promotes self-harm |
| Violence | Incites or instructs violence |
| Child Exploitation | Suggests exploitation, coercion, or blackmail risk involving minors |
| Vulgar Language | Contains insults or profanities |
| Sexual Content | Contains sexual content or solicitation |
| Racist Content | Contains hate, stereotyping, or demeaning of a demographic group |
| Radicalization | Praises, recruits for, or instructs extremist causes |
| Regulated Goods / Illegal Acts | Facilitates illegal goods, services, or criminal activity |
| Education | Promotes academic dishonesty or harmful pressure responses |
| Family | Indicates harmful or unsafe family dynamics |
| Health | Spreads harmful or misleading health advice |

Table 2: Annotation guidelines per harm category.

### 3.6. Quality Assurance

Quality control was applied at two levels. At the *categorical* level, all submitted prompts were reviewed by the project manager to confirm that each entry unambiguously satisfies its assigned category criterion and does not overlap multiple categories. English translations were produced automatically using Google Translate and are included as a reference aid for non-Albanian-speaking readers; they were not systematically post-edited and may contain errors, particularly for Albanian-specific idiomatic expressions that machine translation renders poorly. The final dataset of 2,951 prompts reflects the post-review corpus after removal of duplicates and ambiguous entries identified during quality control.

## 4. Dataset Statistics

Table 3 presents summary statistics for the dataset. Prompts are concise, averaging 8.7 tokens.

| Metric | Value |
|---|---|
| Total Prompts | 2,951 |
| Categories | 11 |
| Language | Albanian (Standard) |
| Average Prompt Length | 8.7 tokens |
| Min Prompt Length | 2 tokens |
| Max Prompt Length | 21 tokens |

Table 3: Dataset statistics.

## 5. Dataset Applications

### 5.1. Primary Use Cases

The dataset is designed to support a range of safety-focused research and development tasks for Albanian. Each use case addresses a distinct stage of the LLM safety pipeline, from evaluation and fine-tuning to deployment and cross-lingual generalization.

1. **Safety Evaluation**: Benchmarking LLM refusal and mitigation behavior on Albanian safety-critical inputs.

2. **Safety Fine-tuning**: Providing supervised training signal for safety-aligned Albanian models.

3. **Red-teaming**: Adversarial probing of deployed systems for Albanian-language vulnerabilities, following the methodology of Ganguli et al. (2022).

4. **Cross-lingual Transfer**: Studying how safety mechanisms trained on other languages generalize to Albanian, motivated by the cross-lingual vulnerability findings of Yong et al. (2023).

5. **Guardrail Development**: Training and evaluating safety classifiers such as Llama Guard (Inan et al., 2023) on Albanian-language content.

### 5.2. Experimental Framework

A standard evaluation protocol using Albanian-LLMSafety proceeds as follows. A target LLM is prompted with each of the 2,951 prompts. Responses are classified as *safe* (refusal or appropriate redirection) or *unsafe* (harmful content generation) either through human annotation or an automated classifier. Per-category and overall refusal rates are then reported, enabling fine-grained diagnosis of model weaknesses across harm types, analogous to the per-category analysis in HarmBench (Mazeika et al., 2024) and SafetyBench (Zhang et al., 2024).

# 6. Ethical Considerations

## 6.1. Ethical Statement

AlbanianLLMSafety contains examples of harmful language spanning violence, exploitation, discrimination, and other categories of harm. These examples exist solely to enable safety research and to help developers identify and mitigate model failures. Their inclusion does not constitute endorsement of any harmful behavior. Access to the dataset is intended for researchers and practitioners working on LLM safety and responsible AI development.

## 6.2. Data Privacy and Responsible Use

Researchers obtaining the dataset are expected to use it exclusively for safety evaluation and model improvement purposes and to comply with applicable ethical guidelines for research involving harmful language data.

## 6.3. Limitations

The dataset has several limitations that should be considered when interpreting results:

- Coverage is necessarily incomplete; new harm categories and evolving attack strategies will require periodic dataset expansion.
- Machine-generated English translations may contain errors, particularly for idiomatic Albanian expressions.
- Prompts are written in Standard Albanian; colloquial register, regional dialectal variation (Gheg, Tosk), and lexical diversity such as synonyms or slang specific to particular communities are not systematically represented. In a safety context, this is a significant gap, as harmful slang and locally relevant harm patterns are often expressed through dialectal and informal varieties rather than Standard Albanian, meaning the benchmark may underestimate safety risks for AI systems deployed across the full range of Albanian-speaking communities.
- Harm category boundaries involve inherent subjectivity, and some prompts may be relevant to multiple categories.

## 6.4. Data Availability

The dataset is released for non commercial research purposes. Users must comply with platform terms of service, avoid attempts at deanonymization, and refrain from surveillance or discriminatory applications. The dataset can be accessed upon request through the following form: `https://forms.gle/YUFdA16R6HkSZjp88`

# 7. Conclusion

We presented AlbanianLLMSafety, a dataset of 2,951 safety evaluation prompts in Albanian covering 11 harm categories. To our knowledge, this is the first dedicated LLM safety benchmark for Albanian, addressing a clear gap in multilingual safety infrastructure. The dataset supports safety evaluation, fine-tuning, red-teaming, and guardrail development for Albanian-language systems, and is positioned to contribute to the broader agenda of multilingual safety research. We hope the dataset will serve as a foundation for both safety evaluation of existing models and the development of safety-aligned systems that better serve Albanian-speaking communities. Future work will expand category coverage, include human evaluation of model outputs against the benchmark, and extend the resource to additional under-resourced language varieties.

# Acknowledgment

This shared task was made possible by the National Priorities Research Program grant NPRP14C-0916-210015 from the Qatar National Research Fund (QNRF), part of the Qatar Research, Development and Innovation Council (QRDI).